\documentclass{article}

\usepackage{microtype}
\usepackage{graphicx}
\usepackage{subfigure}
\usepackage{booktabs} 
\usepackage{tabularx}
\usepackage{listings}
\usepackage{pgfplots}
\usepackage{todonotes}
\usepackage[binary-units=true]{siunitx}

\usepackage{hyperref}

\usepackage[accepted]{icml2019}
\icmltitlerunning{Hijacking Malaria Simulators
	with Probabilistic Programming}

\pgfplotsset{width=7cm,compat=1.8}
\begin{document}
	
	\twocolumn[
	\icmltitle{Hijacking Malaria Simulators
		with Probabilistic Programming}
	
	
	
	

	\icmlsetsymbol{equal}{*}
	
	\begin{icmlauthorlist}
		\icmlauthor{Bradley~J. Gram-Hansen}{equal,en}
		\icmlauthor{Christian Schr\"{o}der de Witt}{equal,en}\\
		\icmlauthor{Tom Rainforth}{st}
		\icmlauthor{Philip H.S. Torr}{en}
		\icmlauthor{Yee Whye Teh}{st}
		\icmlauthor{At\i{}l\i{}m G\"{u}ne\c{s} Baydin}{en}
	\end{icmlauthorlist}

	\icmlcorrespondingauthor{Bradley~J. Gram-Hansen}{bradley@robots.ox.ac.uk}
	
	\icmlaffiliation{en}{Department of Engineering Science, University of Oxford, UK}
	\icmlaffiliation{st}{Department of Statistics, University of Oxford, UK}
	
	\icmlkeywords{Machine Learning, ICML}
	
	\vskip 0.3in
	]
	
	\printAffiliationsAndNotice{\icmlEqualContribution} 
	
	\begin{abstract}
		Epidemiology simulations have become a fundamental tool in the fight
		against the epidemics of various infectious diseases like AIDS and
		malaria.
		However, the complicated and stochastic nature of these simulators
		can mean their output is difficult to interpret, which reduces their usefulness to policymakers.
		In this paper, we introduce an approach that allows one to treat a large class of population-based epidemiology simulators as probabilistic generative models. This is achieved by \emph{hijacking} the internal random number generator calls, through the use of an universal probabilistic programming system (PPS). In contrast to other methods, our approach can be easily retrofitted to simulators written in popular industrial programming frameworks. 
		We demonstrate that our method can be used for interpretable introspection and inference, thus shedding light on black-box simulators. This reinstates much needed trust between policymakers and evidence-based methods. 

	\end{abstract}
	
	\section{Introduction}
	Ending the epidemics of AIDS, tuberculosis, malaria and other infectious diseases by 2030 is a key target within the Good Health \& Well-Being section of the UN Sustainable Development Goals \cite{unitednations,un_sustainable2018}. 
	However, despite decades of substantial international efforts, these diseases kill hundreds of million people a year.
	For example, malaria still annually kills about a quarter of a million children under the age of 5 in Africa alone.
	
	To reach the WHO's target of reducing malaria incidence and mortality rates by at least 90\% by 2030, policymakers are increasingly turning to evidence-based methods, thus oftentimes relying on computational simulations \cite{who_global_2015}. 
	These simulations allow policymakers to infer critical information on disease dynamics and make predictions about the impacts of policies
	before they are rolled out. This frequently increases the effectiveness of interventions and thus ultimately saves resources, or even lives.
	For example, it has been shown that mass vaccination may be largely ineffective in regions of large transmission rates, but may play a crucial role in areas of low transmission \cite{cameron2015defining}. 
	
	Malaria epidemiology is governed by a complex set of drivers, 
	few of which can be understood in isolation \cite{cameron2015defining,autino_epidemiology_2012,smith2008towards,bershteyn2018implementation}.
	These include within-host dynamics, population-specific traits and even local geography.
	Comprehensive modeling of all of these components remains challenging, particularly in a region-specific context. 
	Computational epidemiology simulators have to reflect these complexities and are usually stochastic in nature. This can make simulation output highly non-trivial to interpret, particularly when trying to draw desired inferences coupled with observed data \cite{mwendera_challenges_2019,ferris_openmalaria_2015}.
	
	In this paper, we introduce a novel method that allows one to shed light on the inner workings of a large class of population-based stochastic simulators. We achieve this by extending the work of \citet{baydin2018efficient} by interpreting such population-based simulators as probabilistic generative models within the framework of universal probabilistic programming (UPP) \cite{le-2016-inference}. To this end, we \emph{hijack} existing simulators by overriding their internal random number generators.  Specifically, by replacing the existing low-level random number generator in a simulator with a call to a purpose-built UPP ``controller'', which can thus control, track and manipulate the stochasticity of the simulator.
	
	This allows for a variety of tasks to be performed on
	the hijacked simulator, such as running inference (by conditioning
	the values of certain draws and manipulating others),
	uncovering stochastic structure, and automatically
	producing result summaries, such as establishing the probability
	of different program paths/traces.  By providing a common abstraction 
	framework for different simulators, our approach further allows for
	easy and direct comparison between related or competing
	simulators, a characteristic that is valuable in the context of
	epidemiology simulators \cite{ferris_openmalaria_2015}. We provide a case study of the above in Section \ref{sec:casestudy}.
	
	Our framework already supports application to simulators written in 13 general-purpose programming languages, and is easily extensible. This is crucial as, given the enormous code size and complexity, rewriting epidemiology simulators using a dedicated universal probabilistic programming language, such as Pyro \cite{bingham2019pyro}, is often infeasible.

	In time, we hope our approach will play a critical role in
	bringing recent advancements in probabilistic programming to bear on
	the vast array of existing simulators used throughout the sciences,
	thereby providing wide-ranging impacts across a number of fields.
	
	This paper first gives an overview of existing malaria simulators (Section \ref{sec:background}), and proceed by introducing the necessary background on the \textit{pyprob} framework and the concept of universal probabilistic programming (Section \ref{sec:hijackingsimulators}). Our approach is then demonstrated and analysed in the context of a malaria case study (Section \ref{sec:casestudy}).

	\section{Simulating Diseases}
	\label{sec:background}
	
	In-silico simulators have become a crucial tool in evidence-based decision-making within a large number of disciplines, including statistical physics~\cite{landau_binder_2014}, financial modeling~\cite{jackel2002monte},
	weather prediction~\cite{evensen1994sequential}, epidemiology~\cite{smith2008towards} and many others.
	In many cases, simulation output can augment or even replace real data that may otherwise be costly or even impossible to generate.
	Recent advances in hardware have enabled simulations to model increasingly complex systems.
	Epidemiology studies the prevalence and spreading of diseases across populations. Recent advances in hardware have enabled simulations to model the dynamics of infectious diseases, such as malaria, in ever greater detail.
	
	\subsection{Epidemiology Simulators}
	
	Two the most advanced malaria simulators, namely EMOD~\cite{bershteyn2018implementation} and OpenMalaria  \cite{smith2008towards}, have proven to be particularly valuable to policymakers.
	OpenMalaria is based on microsimulations of \textit{Plasmodium falciparum} in humans and was originally developed to simulate the impacts of malaria vaccines within simple villages or districts.
	Compared with OpenMalaria, EMOD is able to simulate a variety of additional drivers, including complex geographies complete with migration and a large number of policy interventions.
	Both EMOD and OpenMalaria are open source and implemented in C++.

	\section{Hijacking Simulators}
	\label{sec:hijackingsimulators}
	
	Probabilistic programming~\cite{gordon2014probabilistic,staton2016semantics,kozen1979semantics} 
	can be used to express probabilistic models and consequently perform automated inference in these. Once a probabilistic model has been expressed in a probabilistic programming language, a wide range of inference techniques, such as Markov chain Monte Carlo (MCMC) \cite{geyer1992practical}, black-box variational inference~(VI) \cite{ranganath2014black} and amortized inference\cite{le2016inference}, can be used by non-experts in an automated fashion. 
	
	\paragraph{Hijacking a simulator} describes the process by which a simulator's random number generators are replaced by calls to external sampling procedures, which are controlled by a probabilistic programming system (PPS). In practice, this amounts to performing a small number of surgical incisions into the simulator's source code in order to replace built-in calls to random number generators. E.g., given a simulator written in C++/Boost \cite{schaling2011boost}, a Gaussian distribution object \textit{boost::normal} is replaced with the corresponding \textit{pyprob\_cpp} distribution object, namely \textit{pyprob\_cpp::distributions::Normal}. Sampling from this distribution is then done by requesting the PPS to send a sample back to the simulator.
	
	The PPS and the simulator exchange xTensor objects\footnote{\url{https://xtensor.readthedocs.io/en/latest/}} through TCP or IPC using a generic FlatBuffers\footnote{\url{http://google.github.io/flatbuffers/}} protocol. On the PPS side, sampling is done using the deep learning framework PyTorch \cite{paszke2017automatic}.
	
	After a variable has been sampled, using \textit{pyprob\_cpp::sample}, it is sent by the PPS to the simulator, at the same recording the simulator execution trace as a side effect. This allows the PPS to construct sample trace probabilities and other summary statistics (cf. Section \ref{sec:casestudy}). 
	
\begin{figure}
		\centering
		\includegraphics[width=0.45\textwidth]{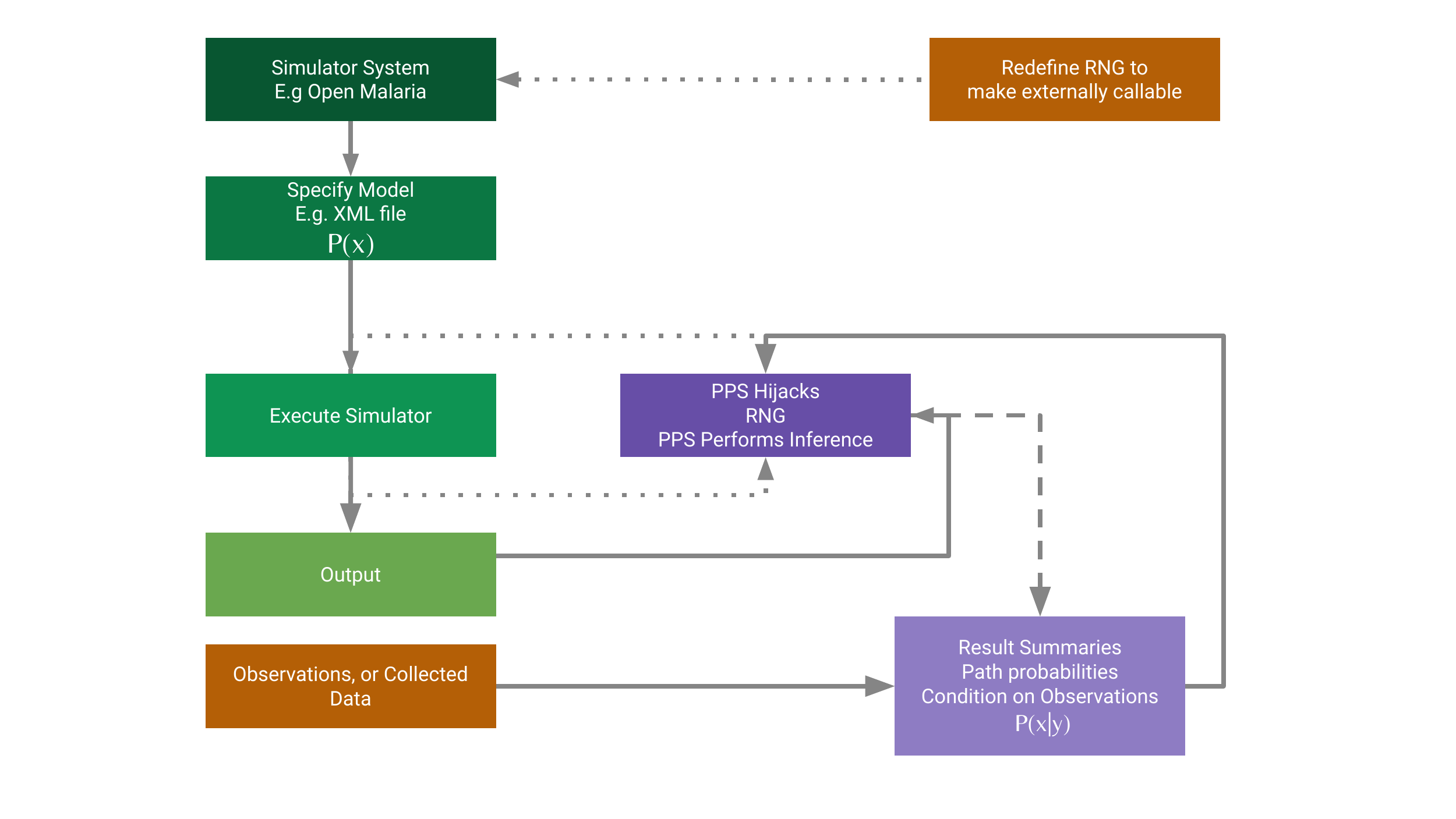}
		\label{fig:how}
		\caption{This flow-chart provides an overview of the process of how our we hijack a generic population-based epidemiology simulator, such as OpenMalaria, and how we modify the simulator to 
			to hijack the random number generator (RNG). It demonstrates how information is exchanged between the simulator and the PPS. Green represents events linked to the simulator, purple corresponds to 
			events occurring in the PPS and brown represents an external processes.}
	\end{figure}

	Finally, the entry point to the simulator, i.e., \textit{main} in C++, is replaced by a special \textit{forward} call, in this case \textit{pyprob\_cpp::forward}.
	This allows the PPS to generate rollouts from the simulator remotely. A pictorial overview of this process is depicted in Figure~\ref{fig:how}.
	
	\paragraph{Population-based simulators} create a trace per population member, in contrast to event-based simulators that create only a single trace per \textit{forward} call~\cite{baydin2018efficient}. This means that, e.g., in OpenMalaria, a standard scenario simulation rollout over the span of $3$ years  
	with a population of size $n=5000$ \cite{smith2008towards} will generate about \textit{four terabytes} of raw trace data. Unlike the simulator used by \citet{baydin2018efficient}, this amount of data cannot feasibly be kept in RAM, which makes \textit{a posteriori} trace analysis very inefficient. 
	In order to deal with the shortcomings of \textit{pyprob} in a population-based simulator context, we therefore extend the framework to be able to do trace analysis on the fly.
	
	By extending \textit{pyprob} to handle population-based simulators, the PPS can now track all stochastic random variables that are created within the simulator, which then allows us to generate trace plots and path probabilities associated to the execution paths of the program.
	The corresponding increase of simulator transparency helps reinstate much needed trust between policymakers and evidence-based methods.

\section{Case Study: Ifakara, Tanzania}
\label{sec:casestudy}

Ensemble methods are commonly used in statistics in order to combine the predictive power of multiple models \cite{cameron2015defining,smith_ensemble_2012}. To this end, recent work has attempted to characterise the similarities and differences between two of the most advanced malaria epidemiology simulators, EMOD \cite{bershteyn2018implementation} and OpenMalaria \cite{smith2008towards}. Evaluation is usually done by comparing a number of output parameters across a range of hand-crafted standard scenarios reflecting different geographical locations across Africa~\cite{smith_ensemble_2012}.

In the following, we illustrate how our method introduces a novel introspection paradigm. By extracting trace graphs from population-based simulators, policymakers can ask specific questions about properties of the model trace flow
and not just the outcomes of the model, thus providing additional interpretability to the decision-making process. 

\begin{figure}
	\centering
	\includegraphics[width=\textwidth/2]{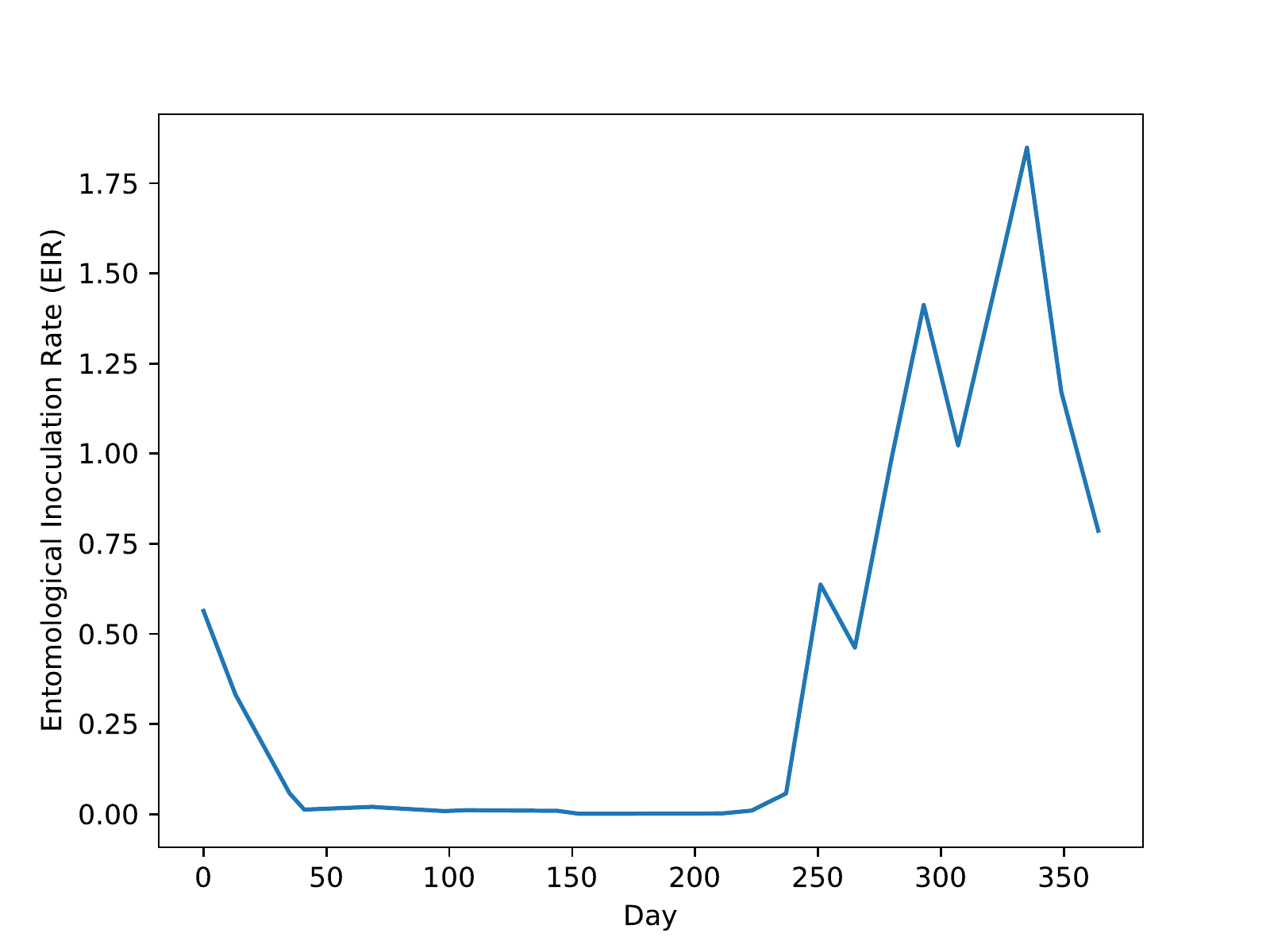}
	\vspace{-20pt}
	\caption{Seasonal Entomological Inoculation Rate (EIR) for the Ifakara scenario. Data is averaged over $30$ day periods.}
	\label{fig:EIR}
	\vspace{-10pt}
\end{figure}

To illustrate the above, we present simulation output generated from a scenario resembling local conditions in the town of Ifakara, Tanzania. Both EMOD and OpenMalaria are configured to simulate a single population node of $n=100$ and assume constant climatic conditions and no migration over the simulation period of three years. Please refer to Figure \ref{fig:EIR} for the seasonal Entomological Inoculation Rate (EIR), a measure of infectivity.

\begin{figure*}[h!]
\centering
\includegraphics[width=\textwidth]{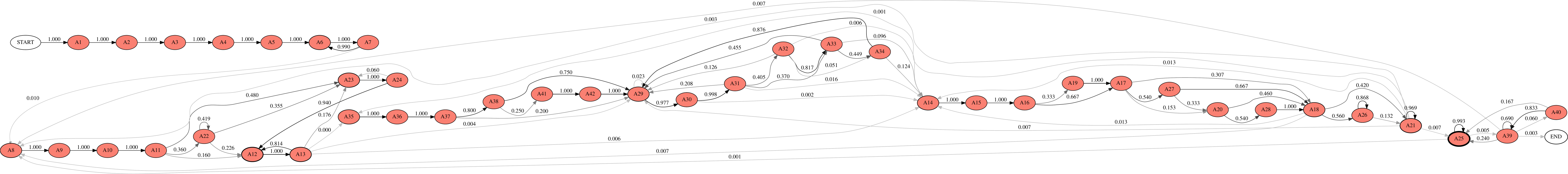}
\includegraphics[width=0.89\textwidth]{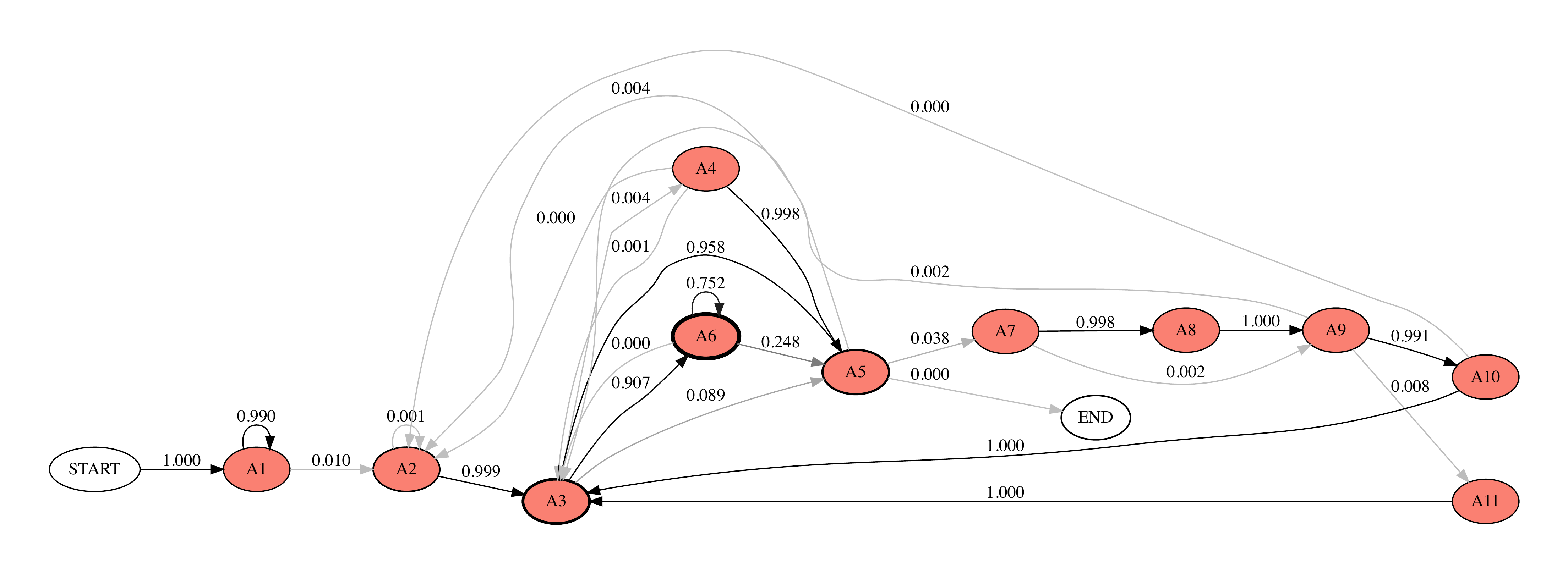}
\caption{Here we run two equivalent models, compare the corresponding trace paths and corresponding 
	path probabilities taken by the thousands of random variables generated internally within the simulators. \textit{\textbf{Top:}} The specified model run in EMOD. \textit{\textbf{Bottom:}} The specified model in OpenMalaria.}
\label{fig:plotewan}
\end{figure*}

We provide examples of the generated trace plots from the connection between the simulator and the PPS in Figure~\ref{fig:plotewan}. 

We can see from the addressing schemes $A_1, \ldots, A_N$ (Tables \ref{table:addresses} and \ref{table:asingleaddress}) what physical events are connected to each other and how outputs in EMOD are generated from a different set of procedures as compared to OpenMalaria.
By having access to such diagrams, users can internally evaluate and scrutinize the decisions that the simulator is making.

This is important for policymakers, or general non-experts, as it not only details how we arrive at the given outputs, but it provides an understanding of which processes were most crucial in determining those outputs as can be seen from the path probabilities assigned to each of the vertices. 

Additionally, by associating nodes in trace graphs representing the same physical processes within different models, the significance of model detail can be evaluated. For example, the full trace presented in Table \ref{table:asingleaddress} represent a sampling step associated with within-host dynamics of the malaria parasite \textit{Falciparum} in OpenMalaria node $A1$. The same physical process also occurs in EMOD's trace graph at position $A7$ (see Section \ref{table:addresses}).
Comparisons like these could help developers better control model complexity, and even provide an alternative testing and debugging paradigm.

\begin{table}[h!]
\footnotesize
\setlength{\tabcolsep}{1mm}
\label{table:asingleaddress}
\caption{An example of an address generated for the model run in the OpenMalaria simulator. We can see that A1
	relates to Generating a member of the human population who may or may not be infect with the malaria disease. We get something similar for EMOD, except this relates to A7 in the EMOD program execution.}
\def\arraystretch{1.25}
\begin{tabularx}{0.46\textwidth}{@{}lX@{}l@{}} 
	\toprule
	Address ID & Full address \\
	\midrule
	A1 & [forward()+0x204; OM::Simulator::
	
	start(scnXml::Monitoring const)+0x28a;
	
	OM::Population::createInitialHumans()+0x94;
	
	OM::Population::newHuman(OM::SimTime)+0x5c;
	
	OM::Host::Human::Human(OM::SimTime)+0x12b;
	
	OM::WithinHost::WHInterface::
	
	createWithinHostModel(double)+0x99;
	
	OM::WithinHost::DescriptiveWithinHostModel
	::DescriptiveWithinHostModel(double)+0x3a;
	
	OM::WithinHost::WHFalciparum::
	
	WHFalciparum(double)+0xe6;
	
	OM::util::random::
	
	gauss(double, double)+0xb4]\_\_Normal \\
	
	$\vdots$ & $\vdots$ \\

		\bottomrule
	\end{tabularx}
\end{table}

\begin{table}[h!]
	\footnotesize
	\setlength{\tabcolsep}{1mm}
	\caption{An interpretation table for each of the address of the 
		overall trace generated from the corresponding OpenMalaria model.}
	\label{table:addresses}
	\begin{tabularx}{0.46\textwidth}{@{}lX@{}} 
		\toprule
		Address ID & Interpretation \\
		\midrule
		A1 & Generate a human in the
		
		population within host dynamics\\

		A2 & Generate another human in
		
		the population within host dynamics\\
		
		A3 & The population is updated and 
		
		a new human, or humans, may get infected \\
		
		A4, A5 & Potential child deaths 
		
		within the population are simulated \\
		
		A6 & Determines parasite density
		
		of an individual infection \\
		
		A7 & Models how the disease is 
		
		progressing within the infected humans \\
		
		A8 & Models how the disease is
		
		progressing within the population \\
		
		A9 & Models how the disease is 
		
		progressing within the infected humans
		
		after the population has been updated\\
		
		A10 & Full clinical update on 
		
		the population for those without severe
		
		or no Malaria infection. \\
		
		A11 & Full clinical update on the population 
		
		for those with severe Malaria infections \\
		\bottomrule
	\end{tabularx}
\end{table}

\section{Discussions and Future Work}

In this work we have demonstrated a method 
that enables one to hijack population-based simulators, extending 
the work of \citet{baydin2018efficient}. 
We applied our method to two malaria-orientated population-based simulators and 
generated a variety of trace graphs.
Finally, we have shown how our system
enables policy makers and non-experts to analyse simulator outputs in a 
way previously unavailable in the field of epidemiology. 

To extend our work further we aim to implement additional 
tools that will facilitate complicated inference procedures that condition on simulator 
output. We will also evaluate additional scenarios across Africa and Southeast Asia to better understand the similarities and differences between EMOD and OpenMalaria.

\section{Acknowledgments}
We thank Ewan Cameron of MAP at the Big Data Institute and the OpenMalaria team at the Swiss Tropical Health Institute for their time and help. BGH is supported by the EPRSC Autonomous  Intelligent Machines and Systems grant. CSW  is supported by the project Free the Drones (FreeD) under the Innovation Fund Denmark and Microsoft. YWT's and TR's research leading to these results has received funding from the European Research Council under the European Union's Seventh Framework Programme (FP7/2007-2013) ERC grant agreement no. 617071. AGB and PH are supported by EPSRC/MURI grant EP/N019474/1 and AGB is also supported by Lawrence Berkeley National Lab.

\bibliographystyle{icml2019}
\bibliography{refs}
\end{document}